\documentclass[journal]{IEEEtran}

\usepackage{cite}
\usepackage{amsmath}
\usepackage{algorithmic}
\usepackage{url}
\usepackage{graphicx}
\usepackage{array}
\usepackage{amssymb}
\usepackage{multirow}
\usepackage[caption=false,font=footnotesize]{subfig}
\usepackage{amsthm}
\usepackage{dsfont}
\usepackage{bm}
\usepackage{booktabs}
\usepackage{xcolor}
\usepackage[T1]{fontenc}
\usepackage{algorithmic}
\usepackage[ruled, vlined]{algorithm2e}
\usepackage{dsfont}

\SetKwInput{KwInput}{Input}             
\SetKwInput{KwOutput}{Output}
\graphicspath{{fig/}}
\def\eg{\emph{e.g.}}

\begin{document}

\title{Semantically Meaningful Class Prototype Learning for One-Shot Image Semantic Segmentation}

\author{
	Tao~Chen,
	Guo-Sen~Xie$^{*}$,
	Yazhou~Yao$^{*}$,
	Qiong~Wang,
	Fumin~Shen,
	Zhenmin~Tang,
	and~Jian~Zhang

\thanks{Tao Chen, Yazhou Yao, Qiong Wang and Zhenmin Tang are with the School of Computer Science and Engineering, Nanjing University of Science and Technology, China.}
\thanks{Guo-Sen~Xie is with the Inception Institute of Artificial Intelligence, Abu Dhabi, United Arab Emirates.}
\thanks{Fumin Shen is with the School of Computer Science and Engineering, University of Electronic Science and Technology of China, China.}	
\thanks{Jian Zhang is with the School of Electrical $\&$ Data Engineering, University of Technology Sydney, Australia.}	
\thanks{*Co-corresponding author.}	
}

\maketitle

\begin{abstract}

One-shot semantic image segmentation aims to segment the object regions for the novel class with only one annotated image. Recent works adopt the episodic training strategy to mimic the expected situation at testing time. However, these existing approaches simulate the test conditions too strictly during the training process, and thus cannot make full use of the given label information. Besides, these approaches mainly focus on the foreground-background target class segmentation setting. They only utilize binary mask labels for training. In this paper, we propose to leverage the multi-class label information during the episodic training. It will encourage the network to generate more semantically meaningful features for each category. After integrating the target class cues into the query features, we then propose a pyramid feature fusion module to mine the fused features for the final classifier. Furthermore, to take more advantage of the support image-mask pair, we propose a self-prototype guidance branch to support image segmentation. It can constrain the network for generating more compact features and a robust prototype for each semantic class. For inference, we propose a fused prototype guidance branch for the segmentation of the query image. Specifically, we leverage the prediction of the query image to extract the pseudo-prototype and combine it with the initial prototype. Then we utilize the fused prototype to guide the final segmentation of the query image. Extensive experiments demonstrate the superiority of our proposed approach. The source codes and models have been made available at \url{https://github.com/NUST-Machine-Intelligence-Laboratory/SMCP}.

\end{abstract}


\ifCLASSOPTIONpeerreview
	\begin{center} \bfseries EDICS Category: 3-BBND \end{center}
\fi
\IEEEpeerreviewmaketitle

\section{Introduction}


\IEEEPARstart{D}{eep} convolutional neural networks (CNNs) have achieved significant breakthroughs in many computer tasks (\eg, image classification \cite{simonyan2014very,yao2020exploiting,zhang2020web,zhang2020web2,li2020field}, image retrieving \cite{yao2019dynamically,hu2020pyretri}, object detection \cite{liu2016ssd}, and semantic segmentation \cite{long2015fully,ronneberger2015u,badrinarayanan2017segnet,chen2020classification,lu2020hsi,luo2019segeqa,zhou2020motion,yu2015multi,chen2017deeplab,fu2019dual,lin2017refinenet}, video streaming \cite{gao2018optimizing,zhan2019unmanned}). However, training deep CNNs typically requires large-scale labeled datasets, which is expensive to obtain. Though semi-supervised \cite{li2020semi}, weakly-supervised \cite{jiang2019integral,liu2021exploiting,yao2020bridging,sun2020crssc,zhang2020data}, and unsupervised methods \cite{yao2019towards,zou2018unsupervised,yao2017exploiting,xie2019attentive,xie2020region,yao2017new,zhang2020deep,yao2016domain,wang2020set,yao2016automatic} are recently proposed to alleviate the annotation burden, these traditional deep CNNs are trained for predefined classes. Consequently, they cannot generalize well to the tasks with new incremental categories that are not defined during training. Even if given several samples of the novel classes, the trained network is still hard to be fine-tuned in a data-efficient way \cite{ding2020approximate,yao2018discovering,sun2020exploiting,yao2018extracting}. In contrast, with the learning of only one annotated image containing the new class, humans can successfully recognize this new category and segment its regions. To imitate such a generalization ability of human beings, researchers recently move their attention to the task of one-shot learning, which can also reduce the data-gathering effort. In this paper, we focus on solving the one-shot learning problem for semantic segmentation. In our setting, each test image contains one new target category, and our task of one-shot image segmentation \cite{shaban2017one,rakelly2018conditional,dong2018few,hu2019attention,zhang2020sg,wang2019panet,siam2019amp,zhang2019canet} is to predict the object regions for that unseen semantic class with only one annotated image for that new class.

Here, we first define the terms for "support image", "support mask", "query image", "pseudo-mask", "support prototype" and "pseudo-prototype". We refer to the image-mask pair that provides clues for the new class as the support set (the image and mask are denoted as the support image and support mask, respectively). Meanwhile, we refer to the image that needs to be segmented as the query image. During testing, we denote the initial prediction of the query image as the pseudo-mask. The class prototype obtained with the support image-mask pair is denoted as the support prototype. In contrast, the prototype obtained with the query image and its pseudo-mask is denoted as the pseudo-prototype.

\begin{figure}[t]
	\centering
	\includegraphics[width=0.9\linewidth]{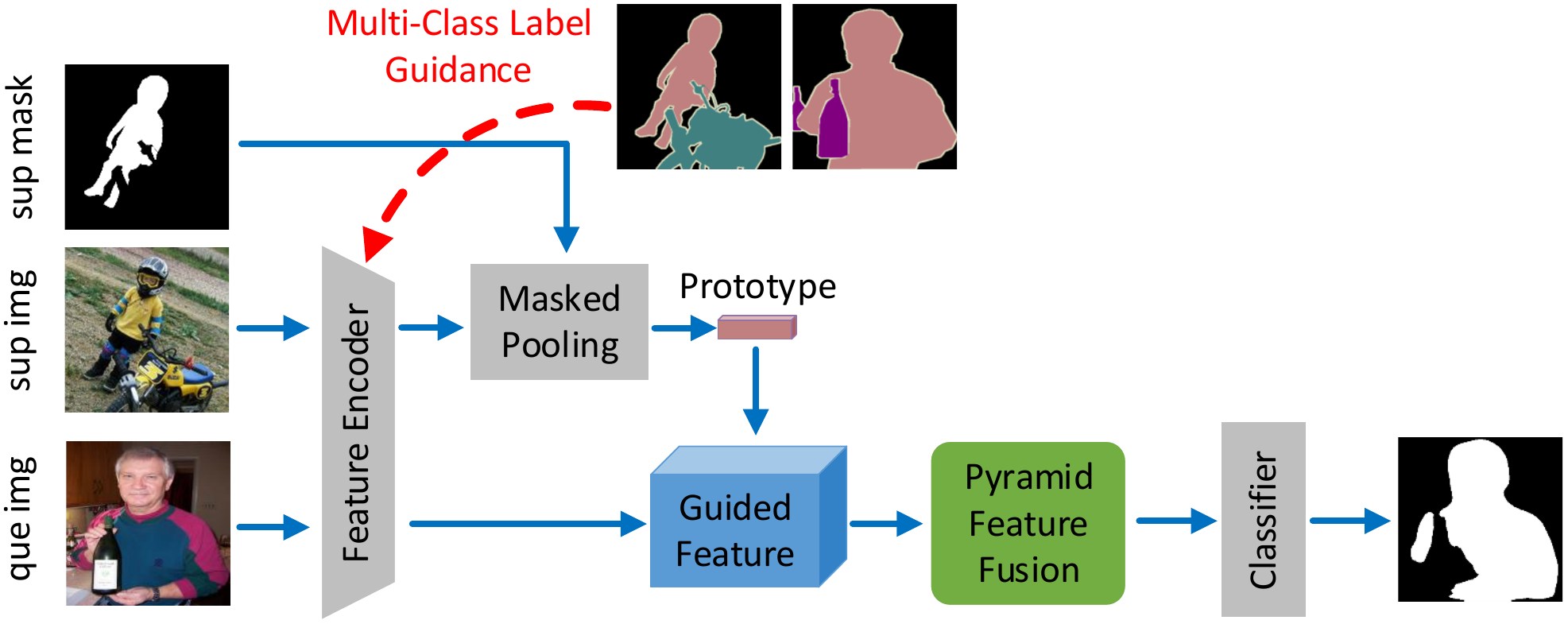}
	\caption{Overview of our framework. Different from existing works, we propose to leverage the multi-class label information during the episodic training. After obtaining the guided features, we propose a pyramid feature fusion module to mine the corresponding features for the target class in a multi-scale way.}
	\label{fig_moti}
\end{figure}

Humans' capability of learning with only limited examples for new tasks is, to some extent, based on their past experiences. It is thus crucial for one-shot learning to leverage the prior knowledge, for example, a large amount of annotated seen class images. Since fine-tuning a traditional semantic segmentation network on a single labeled image for the new category is prone to over-fitting, recent works adopt an episodic strategy when training the network on a dataset that can provide the prior knowledge. For the episodic training, the setting is constructed to mimic the expected situation at the testing time. When conditioned on a single support example with the annotation of the unseen class, the network trained in such a meta-learning way aims to perform segmentation well for the target class in the query image. However, these existing approaches simulate the test conditions too strictly during the training process. They cannot make full use of the given label information. For example, these approaches mainly focus on the foreground-background target class segmentation setting. They only utilize binary mask labels for training and discard the multi-class label information. Without the multi-class label information, the features extracted from the encoder will lack semantic information. The network is also more prone to over-fitting, which hinders the design of subsequent fusion networks for the guided features. Besides, in the existing training strategy for one-shot segmentation, the support image-mask pair is only used in the conditional branch to provide guidance information for the segmentation branch, which can be further utilized to facilitate the network training. 

To tackle the above challenges, as illustrated in Fig.~\ref{fig_moti}, in this work, we first propose to take advantage of the multi-class label information during the episodic training. This will encourage the feature encoder to generate more semantically meaningful feature representations for both support and query images. The class-aware semantic features of the support branch will then lead to a semantically meaningful class prototype for the target class. We then further propose a pyramid feature fusion module to integrate the target class cues obtained from the support branch into the query features. Our pyramid feature fusion module can mine the corresponding features for the target class in a multi-scale way to generate class-agnostic but robust features for the final atrous spatial pyramid pooling (ASPP) \cite{chen2017deeplab} classifier. That is, on the one hand, our feature encoder focuses on extracting class-aware semantic feature representations from the input images. On the other hand, the pyramid feature fusion module aims to exploit the fused information and generate class-agnostic features for the binary prediction of the target category. 

Additionally, to take more advantage of the support image-mask pair, we propose a self-prototype guidance branch for the support image segmentation. After obtaining the target class prototype with the support image-mask pair, as illustrated in Fig.~\ref{fig_f}, we not only use it to guide the segmentation of the query image but also leverage the class prototype to guide the segmentation of the support image. Our proposed training scheme will constrain the network to generate more compact features and robust prototype for each semantic class. This will help the network to locate the corresponding regions of the target category more accurately. Finally, we propose a fused prototype guidance branch at the test time. It will help obtain a more robust target class prototype for the segmentation of the query image. Since the label of the query image is not available in the test phase, we take the predicted binary segmentation map as the pseudo-mask to extract the pseudo-prototype for the query branch. Then we fuse the class prototype generated from the support image and pseudo-prototype from the query image. Then we utilize the fused prototype to guide the final segmentation of the query image.
The main contributions of this work are: 

\textbf{1)} We propose to leverage the multi-class label information to constrain the network during the episodic training of one-shot semantic image segmentation. This will help the network extract class-aware semantic feature representations from the input images and generate a more semantically meaningful class prototype for the target category.

\textbf{2)} To take advantage of the support image-mask pair, we propose a self-prototype guidance branch for the support image segmentation. This will encourage the network to learn more compact features and a more robust target class prototype, and thus can help to locate the corresponding area of the target class more accurately.

\textbf{3}) During testing, we propose a fused prototype guidance branch to leverage the target class prototypes from both support and query images to guide the final segmentation of the query image.

\textbf{4)} Extensive experiments on the PASCAL-$5^{i}$ and COCO-$20^{i}$ datasets demonstrate the superiority of our proposed approach for one-shot image segmentation.

The rest of this paper is organized as follows: 
the related work is described in Section \ref{related_work} and our approach is introduced in Section \ref{approach}; we then report our evaluations on two widely-used datasets for one-shot image segmentation task in Section \ref{experiments}; we report the ablation studies in Section \ref{ab_study} and finally conclude our work in Section \ref{conclusion}.

\section{Related Work}
\label{related_work}

\subsection{Semantic Segmentation}

Different from image classification that only needs to attach one label to an input image, semantic segmentation \cite{long2015fully,ronneberger2015u,badrinarayanan2017segnet,yu2015multi,chen2017deeplab,fu2019dual} aims to label every pixel of the input image. After the introduction of the fully convolutional network (FCN) \cite{long2015fully}, deep learning has achieved great success in semantic segmentation. For example, the early work of UNet \cite{ronneberger2015u} and SegNet \cite{badrinarayanan2017segnet} used encoder-decoder architecture to recover the spatial resolution of the input image, for medical image segmentation and street scene parsing, respectively. Then dilated/atrous convolution was proposed in \cite{yu2015multi, chen2017deeplab} to enlarge the receptive field of the network while maintaining the spatial resolution of the feature maps. To produce high-resolution segmentation maps, RefineNet \cite{lin2017refinenet} leveraged a cascaded architecture to combine low-resolution semantic features and fine-grained low-level features in a recursive manner. Recently, contextual information for semantic segmentation was explored in \cite{zhang2018context} to capture the global semantic context of scenes, and selectively highlight class-dependent features. To capture long-range dependencies, the works of \cite{huang2019ccnet,wang2018non,fu2019dual} leveraged non-local neural networks for self-attention to adaptively integrate local features with their global dependencies. A joint multi-task learning framework for semantic segmentation and boundary detection \cite{zhen2020joint} was proposed. In their work, an iterative pyramid context module was adopted to couple the two tasks and store the shared latent semantics to interact between semantic segmentation and boundary detection. Dynamic routing for semantic segmentation \cite{li2020learning} was proposed to alleviate the scale variance. They generated data-dependent routes for adapting to the scale distribution of each image.

\subsection{One-Shot Classification}
One-shot classification algorithms \cite{finn2017model, munkhdalai2017meta, snell2017prototypical, sung2018learning,wu2019parn,peng2019few,wang2020generalizing} aim to learn information about object categories with only one training example for each category. Due to the rareness of the sample, recent works choose to generalize knowledge acquired from seen classes during training to new categories rather than directly using supervised learning-based approaches. For example, model-agnostic meta-learning \cite{finn2017model} was compatible with any gradient-based training model. It can generalize well with only a small number of training samples. MetaNet \cite{munkhdalai2017meta} acquired knowledge in a meta-learning way. It transferred its parameters and biases via fast parameterization for the new tasks. Recently, metric-based approaches were proposed to learn a metric space for classification during meta-learning. Prototypical networks \cite{snell2017prototypical} proposed to classify images of new classes by computing distances to prototype representations of each class. Relation network \cite{sung2018learning} proposed to classify images by computing the relation scores between the query and the support image of each new category. To address the high-variance issue, the ensemble of deep networks was designed in  \cite{dvornik2019diversity} to encourage the networks to cooperate. The work of \cite{peng2019few} jointly incorporated visual feature learning, knowledge inferring, and classifier learning into one unified framework for knowledge transfer. DeepEMD \cite{zhang2020deepemd} formalized the one-shot classification as an optimal matching problem between image regions. Earth Mover’s Distance was adopted in their work as the distance metric between the structured representations to determine image relevance.

\subsection{One-Shot Semantic Segmentation}
One-shot semantic image segmentation is the task of segmenting the object pixels with only one annotated image \cite{shaban2017one,rakelly2018conditional,zhang2019canet,hu2019attention,zhang2020sg,dong2018few,wang2019panet,wang2017multi,wang2020lt}. The target class is defined by the binary ground-truth segmentation mask of the support image. Shaban et al. proposed the first model OSLSM \cite{shaban2017one} for one-shot semantic segmentation with a two-branched architecture. In their approach, a conditioning branch was adopted to analyze the target class in the support image. Then parameters were generated for the query features to perform segmentation. Recent progress for one-shot segmentation mainly followed such two-branched architecture. For example, co-FCN \cite{rakelly2018conditional} and CANet \cite{zhang2019canet} extended OSLSM \cite{shaban2017one} by leveraging the support branch to extract an encoded feature embedding. Then the feature representation that contains the information of the target class was later fused to the query branch as guided information. A-MCG \cite{hu2019attention} proposed an attention-based multi-context guiding network to integrate multi-scale context features between support and query branches. And spatial attention along the fusion branch was adopted in their work to enhance self-supervision in one-shot learning.  Another extended version was the similarity guidance network SG-One \cite{zhang2020sg}. The similarity between the masked pooled support feature and feature maps of the query image was calculated to guide the segmentation of the query branch. While the above methods adopt a parametric module to fuse the support information with query features, PL \cite{dong2018few} and PANet \cite{wang2019panet} proposed to leverage non-parametric metric learning to solve the segmentation task. They proposed to extract class-specific prototype representations within an embedding space. Segmentation was then performed over the query images by matching each pixel to the learned prototypes. The key to the success of these methods lies in the effective utilization of information from the support image to assist the segmentation of the query image. They leverage metric learning to perform feature matching \cite{shaban2017one,rakelly2018conditional,zhang2019canet,hu2019attention,dong2018few} (or calculate a certain distance \cite{zhang2020sg,wang2019panet}, such as cosine similarity) to mine the connection between the support and query images, thereby improving the query image's segmentation results. However, existing methods only utilize binary mask labels for training and discard the multi-class label information. Therefore, features extracted from the encoder will be less semantically meaningful, which hinders the design of subsequent fusion networks for the guided features.

\begin{figure*}[t]
	\centering
	\includegraphics[width=0.98\textwidth]{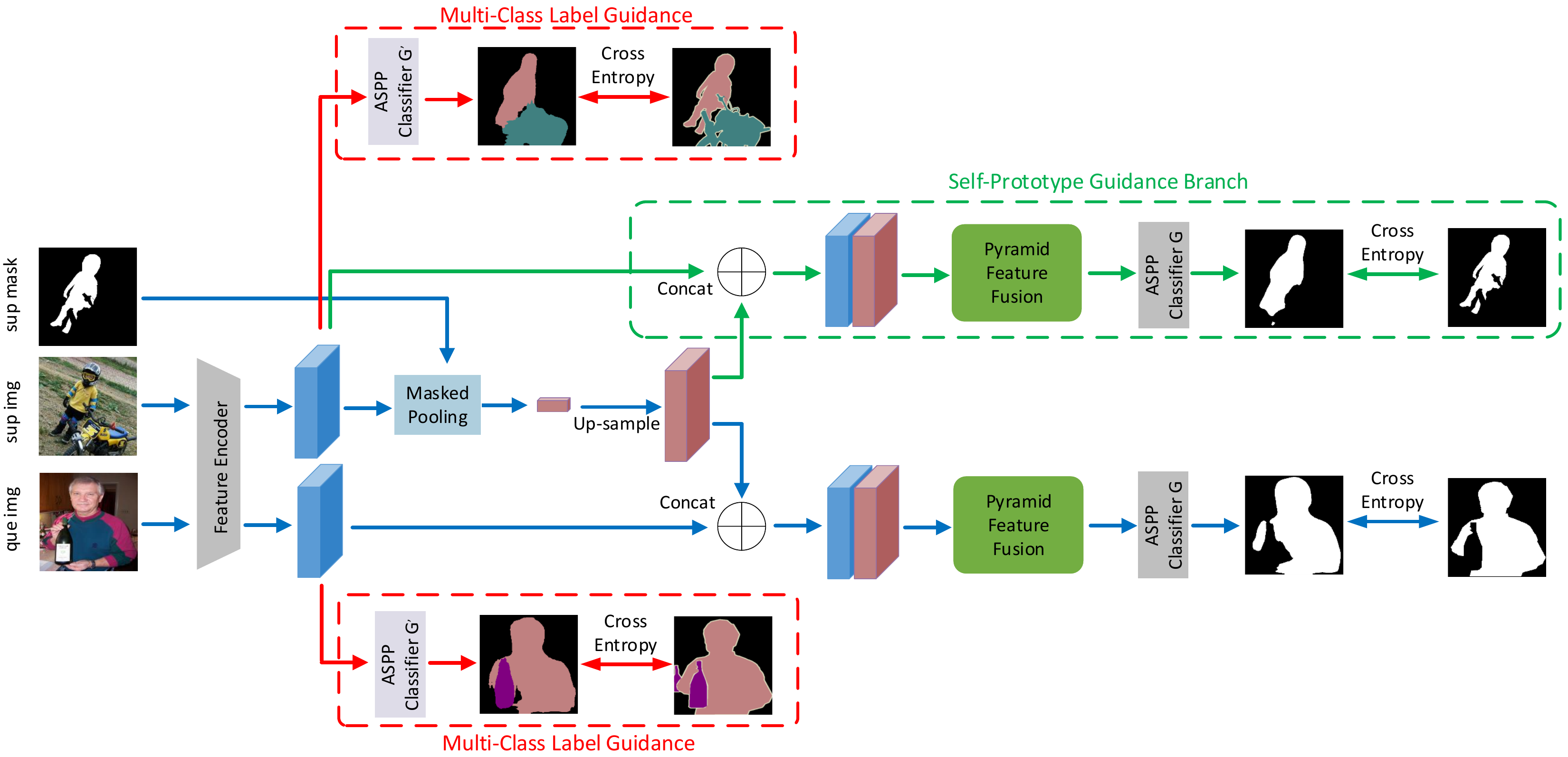}
	\caption{Illustration of our training framework. For each episode, a feature encoder is first used to extract deep features for the support and query images. On the one hand, the support and query features are forwarded to the multi-class label guidance branch for traditional semantic segmentation training. On the other hand, masked average pooling is applied with the support mask to obtain the target class feature prototype from the deep support feature maps. The target class prototype vector is then up-sampled and concatenated to both the support and query features for the binary prediction of the target category. Best viewed in color.}
	\label{fig_f}
\end{figure*}

\section{The Proposed Approach}
\label{approach}

\subsection{Problem Setting}

Our task is to learn a segmentation model, which can predict the pixels of the unseen semantic class $c_{j} \in C_{test}$ in the query image, given only one support image with a corresponding binary mask for $c_{j}$ during testing. Note that we follow the early works \cite{shaban2017one,rakelly2018conditional,dong2018few,hu2019attention,zhang2020sg,wang2019panet} and focus on solving the one-way setting, which means each image contains only one target class. During training, we can access a large set of images with pixel-level ground-truth labels. For each training image $\bm{I} \in \mathds{R}^{h \times w \times 3}$, we denote its semantic label as $\bm{Y} \in \mathds{R}^{h \times w \times c}$, where (h, w) is the size of the image and c is the number of classes. For one-shot semantic image segmentation setting, $C_{train}$ and $C_{test}$ are two non-overlapping sets of classes ($C_{train}\cap C_{test}= \varnothing$), which is different from the traditional image segmentation task. In other words,  the class $c_{j}$ used during testing is never labeled during the training process. 

We take the widely adopted episodic training strategy to match the expected situation at testing time. Each training episode instantiates a one-shot semantic image segmentation task for a class $c_{i}\in C_{train}$. Specifically, two images containing class $c_{i}$ are selected as a pair of support-query images. Similarly, each testing episode instantiates an one-shot task for a class $c_{j}\in C_{test}$. Two images containing class $c_{j}$ are selected as a pair of support-query images. Unlike existing one-shot image segmentation approaches \cite{shaban2017one,rakelly2018conditional,dong2018few,hu2019attention,zhang2020sg,wang2019panet} that only leverage the binary mask $\bm{M_{i}}$ for class $c_{i}$, we also take advantage of the pixel-level ground-truth labels $\bm{Y}$ during training like the traditional semantic segmentation.

\subsection{Base Network}
\label{sec_base_net}

Fig.~\ref{fig_f} shows the architecture of our episodic training scheme. For each episode, the base network first uses a feature encoder to extract deep features for the support and query images. Then the masked average pooling \cite{zhang2020sg} is applied with the support mask to obtain the prototype vector for the target class from the support feature maps. We up-sample the support features \bm{$F_{s}$} to the same size of the mask $\bm{M_{s}}$ and then the prototype $\bm{p}$ is calculated as:
\begin{equation}
	\bm{p}=\frac{\sum\limits_{h,w} \bm{M_{s}}^{(h,w)} \cdot \bm{F_{s}}^{(h,w)}}{\sum\limits_{h,w} \bm{M_{s}}^{(h,w)}}  ,
\end{equation}
where $(h, w)$ is the size of the input image and mask. We then use this prototype containing the class information to guide the segmentation of the query image. We up-sample the prototype vector to the same spatial size of query features and concatenate it to the query features. For the base network, we leverage one convolutional layer with the kernel size of $3\times3$ to fuse the concatenated features. This convolutional layer will be replaced by our proposed pyramid feature fusion module after we introduce the multi-class label information. Then a foreground and background ASPP classifier $G$ is applied on top of the fused features $\bm{F_{f}}$ to get the segmentation map of the query image. The one-shot segmentation loss for the query image is defined as the cross-entropy loss between the prediction $G\left (\bm{F_{f}} \right)$ and query mask $\bm{M_{q}}$:
\begin{equation}
	{L}_{q}= - \sum_{h,w}\sum_{c\in C}\bm{M_{q}}^{\left (h,w,c \right )}
		\log \left ( G\left (\bm{F_{f}} \right )^{\left ( h,w,c \right )} \right ) 
	\label{eq_fg}.
\end{equation}
Here, $C = \left \{ 0,1 \right \}$ is the class label that denotes whether the pixel belongs to the target class. $(h, w)$ is the size of the input query image and mask.

\subsection{Multi-Class Label Guidance}

For traditional semantic image segmentation, pixel-level multi-class labels are provided to train the network. When it comes to the one-shot task, recent works only leverage the foreground and background binary mask for the target class. They intend to meta-learn a class agnostic segmentation network. However, as pointed out in OSLSM \cite{shaban2017one}, part of the recent algorithms' ability to generalize well to unseen classes is benefiting from the pre-training performed on ImageNet \cite{deng2009imagenet}. The weak image-level annotations for a large number of categories endow the pre-trained network with the particular ability to extract discriminative features for both seen classes during training and unseen classes during testing. The existing approaches without any label information during training will gradually result in less semantically meaningful features. Therefore, we propose to leverage the pixel-level multi-class labels to constrain the feature extraction of the encoder. This will encourage the encoder to generate more discriminative features for each category. The more semantically meaningful features will lead to the target class feature prototype with more robust class information. Thus, the prototype will help the network locate the area for the target class more accurately. As shown in Fig.~\ref{fig_f}, we apply a parameter-shared multi-class classifier $G{}'$ on top of the features $\bm{F}$ of both support and query images for traditional multi-class semantic image segmentation. The multi-class segmentation loss is defined as the cross-entropy loss between the prediction $G{} '\left ( \bm{F} \right )$ and multi-class label $\bm{Y}$:
\begin{equation}
	{L}_{seg}= - \sum_{h,w}\sum_{c\in C}\bm{Y}^{\left (h,w,c \right )}
		\log \left ( G{}'\left ( \bm{F} \right )^{\left ( h,w,c \right )} \right ) 
	\label{eq_multi}.
\end{equation}
Here, $C = \left \{ 0, 1, ... , |C_{train}|-1 \right \}$ and $(h, w)$ is the size of the input image and mask. With the multi-class label guidance, our feature encoder focuses on extracting class-aware semantic feature representations. Our pyramid feature fusion module will then process the fused features and generate class-agnostic features for the binary prediction of the target category.

\begin{figure}[b]
	\centering
	\includegraphics[width=\linewidth]{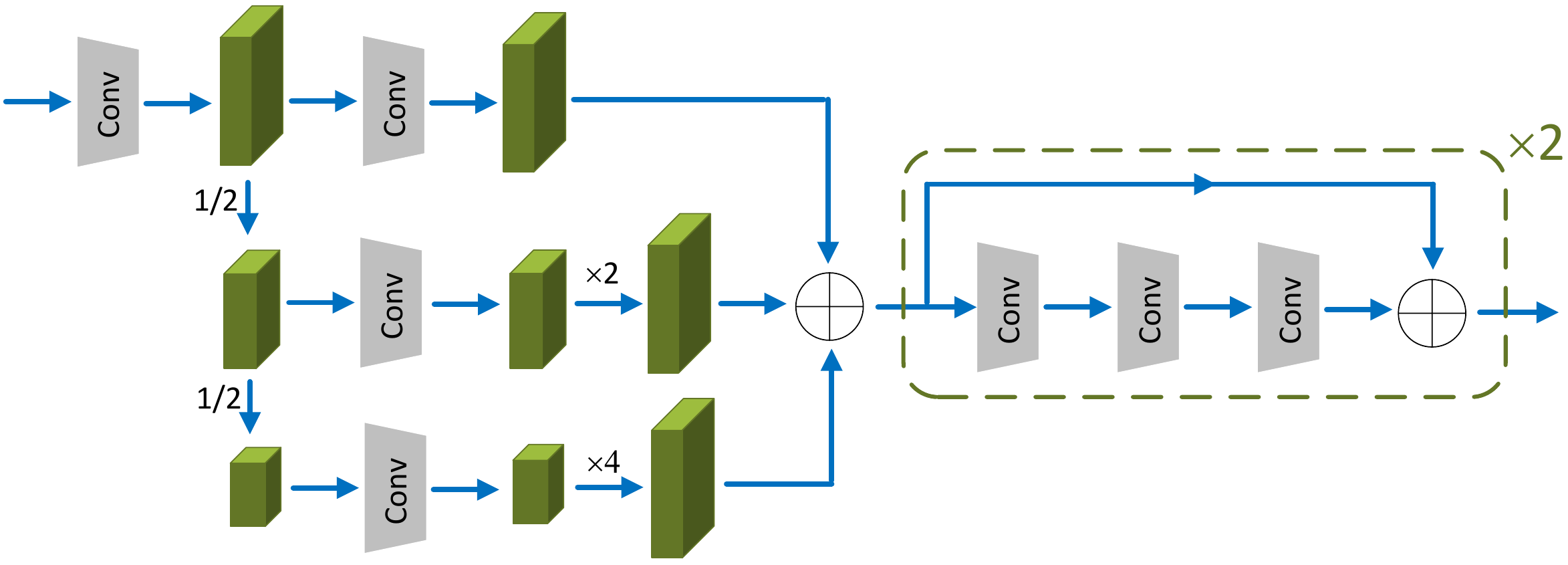}
	\caption{The architecture of our proposed pyramid feature fusion module.}
	\label{fig_pff}
\end{figure}

\begin{figure*}[t]
	\centering
	\includegraphics[width=0.98\textwidth]{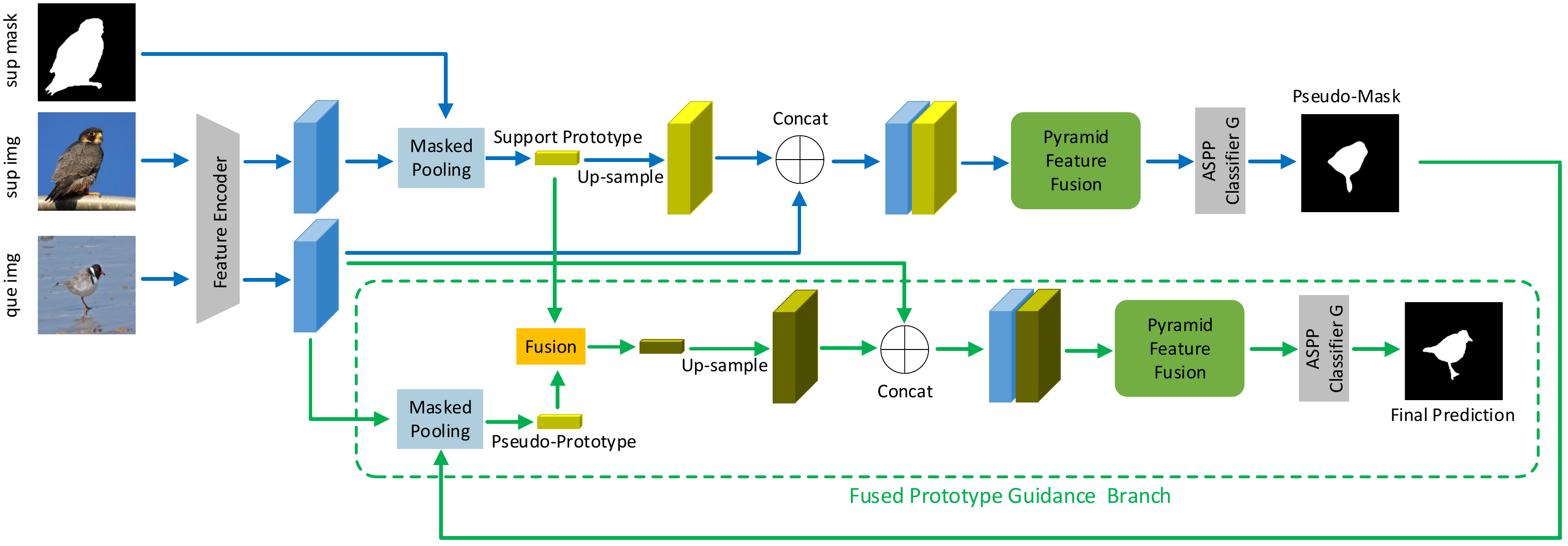}
	\caption{Illustration of the testing framework for our proposed approach. After obtaining the segmentation map for the query image, we treat it as the pseudo-mask to extract the pseudo-prototype for the target class. Then we fuse the pseudo-prototype with the support prototype to guide the final segmentation of the query image. Best viewed in color.}
	\label{fig_t}
\end{figure*}

\subsection{Pyramid Feature Fusion Module}

Given more discriminative features and semantically meaningful guidance information for the target class, we propose a pyramid feature fusion module to better integrate the segmentation cues into query features. The architecture is shown in Fig.~\ref{fig_pff}. The module's input is the concatenation of query feature maps and the up-sampled support prototype. We first apply a $3\times3$ convolutional layer to reduce the dimension of the concatenated features from 1024 to 512. We then down-sample feature maps to 1/2 and 1/4 of the original spatial size. After that, we forward features of each size into a $3\times3$ convolutional layer with 512 filters separately to mine the features in a multi-scale way. We then up-sample the features to their original size before adding them together element-wisely. Finally, we add two residual blocks to enhance the fused features further and to obtain the class-agnostic features for the binary prediction of the target category. Each residual block contains three convolutional layers with 64, 64, and 512 filters separately. Our above pyramid feature fusion module can mine the corresponding features for the target class to generate features robust to the object scales for the final ASPP classifier. Compared with pyramidal feature hierarchy methods \cite{liu2016ssd,lin2017refinenet} which require multi-stage features, our three paths for down/up-sampling pyramidal fusion construct the different resolution representations at a single stage. Compared to ASPP that enlarges the receptive field while maintaining feature resolution, our pyramidal fusion aims to generate various resolution features for both coarsely and finely locating the target class's areas. Note that, we combine our pyramid feature fusion module and ASPP for feature fusion module and classification module, respectively, to construct the segmentation head.

\subsection{Self-Prototype Training Branch}

In existing approaches, the network learned during training is directly applied to the one-shot segmentation task for testing without fine-tuning. Therefore, during testing, the support image-mask pair is only used to provide guidance information about the target class for the segmentation of the query image. Nevertheless, the support image-mask pair can be further utilized during training. To generate a robust class prototype that is informative to guide the network for locating the corresponding target area, we propose a self-prototype guidance branch at the time of training. We expect that the target class prototype generated from the conditioning branch can effectively guide the segmentation of the support image itself. On top of the fused features $\bm{F_{f{}'}}$ of support feature maps and the up-sampled support prototype, we apply the same query image segmentation head to get the binary segmentation map for the support image. The benefits of the self-prototype guidance branch are threefold. First, it provides more supervision for the segmentation head applied on top of fused features. This can alleviate the confusion of the segmentation head when the differences between the support and query features are too significant. Second, the self-prototype guidance branch ensures that the class prototype extracted from support features can effectively help locate the target class in the support image in turn. This constrains the network to generate more compact features and robust prototype for each semantic class. Third, it also echoes with the fused prototype guidance branch during testing (in Section~\ref{sec_pf}), which leverages the self-prototype (pseudo-prototype) of the query image to guide the final segmentation of the query image. Similar to Equation~\eqref{eq_fg}, the one-shot segmentation loss for the support image is defined as the cross-entropy loss between the prediction $G\left (\bm{F_{f{}'}} \right )$ and support mask $\bm{M_{s}}$: 
\begin{equation}
	{L}_{s}= - \sum_{h,w}\sum_{c\in C}\bm{M_{s}}^{\left (h,w,c \right )}
		\log \left ( G\left (\bm{F_{f{}'}} \right )^{\left ( h,w,c \right )} \right ). 
\end{equation}
Here, $C = \left \{ 0,1 \right \}$ is the class label that denotes whether the pixel belongs to the target class. $(h, w)$ is the size of the input support image and mask.

\subsection{Overall Training Objective}

The overall training objective is to learn a one-shot image segmentation network. At the same time, we leverage the multi-class label information to encourage the feature encoder for learning semantically meaningful features for each category. The cost function is as follows:
\begin{equation}
	 L = {L}_{q} + {L}_{s} 
		+ \lambda _{mcl}{L}_{seg}.
\end{equation}
Here, $\lambda _{mcl}$ is the hyper-parameter that controls the relative importance of the one-shot image segmentation loss and the traditional semantic segmentation loss of the multi-class label guidance branch.

\subsection{Testing Architecture with Prototype Fusion}
\label{sec_pf}

Despite our efforts to learn class-aware semantic features and extract the semantically meaningful prototype for the target class during training, the visual appearance and layout differences between the support and query images will make their features to be more or less different. Therefore, at the test time, we propose a fused prototype guidance branch to guide the segmentation of the query image with more robust class cues. We leverage the network to predict the segmentation map for the query image. Then we treat the predicted binary map as the pseudo-mask $\bm{M_{q}}$ to extract the pseudo-prototype for the target class. The masked average pooling \cite{zhang2020sg} is applied with the pseudo-mask to obtain the prototype vector for the target class from the query feature maps. We up-sample the query features $\bm{F_{q}}$ to the same size of the pseudo-mask $\bm{M_{q}}$ and then the prototype is calculated as:
\begin{equation}
\bm{p{}'}=\frac{\sum\limits_{h,w} \bm{M_{q}}^{(h,w)} \cdot \bm{F_{q}}^{(h,w)}}{\sum\limits_{h,w} \bm{M_{q}}^{(h,w)}}  ,
\end{equation}
where $(h, w)$ is the size of the query image and pseudo-mask. Benefiting from the self-prototype guidance branch during training, the network can be directly applied to the pseudo-prototype (self-prototype for the query image) setting. However, the pseudo-prototype of the query image might be noisy due to the coarseness of the pseudo-mask. As illustrated in Fig.~\ref{fig_t}, we then further fuse the pseudo-prototype of the query image with the support prototype to guide the final segmentation of the query image. Here, we combine them by averaging the two prototypes. Our experiments in Section~\ref{subsec_ab} prove that utilizing the fused prototype can achieve better performance than using either the query pseudo-prototype or the support prototype only.

\begin{table*}[t]
	\setlength{\tabcolsep}{7mm}
	\renewcommand{\arraystretch}{1.3}
	\caption{Results of one-shot segmentation on PASCAL-$5^{i}$ dataset evaluated with mean-IoU metric.}
	\label{tab_voc_m}
	\centering
	\begin{tabular}{rccccc}
		\toprule
		\textbf{Method\:\:\:}  &\textbf{PASCAL-}\bm{$5^{1}$}&\textbf{PASCAL-}\bm{$5^{2}$}&\textbf{PASCAL-}\bm{$5^{3}$}&\textbf{PASCAL-}\bm{$5^{4}$}&\textbf{Mean}\\
		\hline
		LogReg \cite{shaban2017one}&26.9&42.9&37.1&18.4&31.4\\
		Siamese \cite{shaban2017one}&28.1&39.9&31.8&25.8&31.4\\
		1-NN \cite{shaban2017one} &25.3&44.9&41.7&18.4&32.6\\
		OSLSM \cite{shaban2017one}   & 33.6&55.3&40.9&33.5&40.8 \\
		co-FCN \cite{rakelly2018conditional} & 36.7&50.6&44.9&32.4&41.1\\
		AMP \cite{siam2019amp}       & 41.9&50.2&46.7&34.7&43.4\\
		SG-One \cite{zhang2020sg}   & 40.2&58.4&48.4&38.4&46.3\\
		PANet \cite{wang2019panet}     & 42.3&58.0&\textbf{51.1}&41.2&48.1\\
		\hline
		\textbf{Ours}  &\textbf{50.6}&\textbf{61.9}&49.4&\textbf{48.4}&\textbf{52.6}\\
		\bottomrule
	\end{tabular}
\end{table*}

\begin{table}[t]
	\renewcommand{\arraystretch}{1.3}
	\caption{Results of one-shot segmentation on PASCAL-$5^{i}$ dataset evaluated with the binary-IoU metric. The experiments are conducted on 4 splits and the average results are reported.}
	\label{tab_voc_b}
	\setlength{\tabcolsep}{6mm}
	\centering
	\begin{tabular}{rc}
		\toprule
		\textbf{Method\:\:\:\:}  &\textbf{binary-IoU}\\
		
		\hline		
		
		FG-BG \cite{rakelly2018conditional}       & 55.0 \\
		Fine-tuning \cite{rakelly2018conditional} & 55.1 \\
		co-FCN \cite{rakelly2018conditional}      & 60.1\\
		OSLSM \cite{shaban2017one}        & 61.3 \\
		PL \cite{dong2018few}             & 61.2 \\
		A-MCG \cite{hu2019attention}            & 61.2\\
		AMP \cite{siam2019amp}            & 62.2\\
		SG-One \cite{zhang2020sg}        & 63.9\\
		PANet \cite{wang2019panet}          & 66.5\\
		\hline
		\textbf{Ours}  &\textbf{68.7}\\
		\bottomrule
	\end{tabular}
\end{table}

\begin{figure*}[t]
	\centering
	\includegraphics[width=0.99\textwidth]{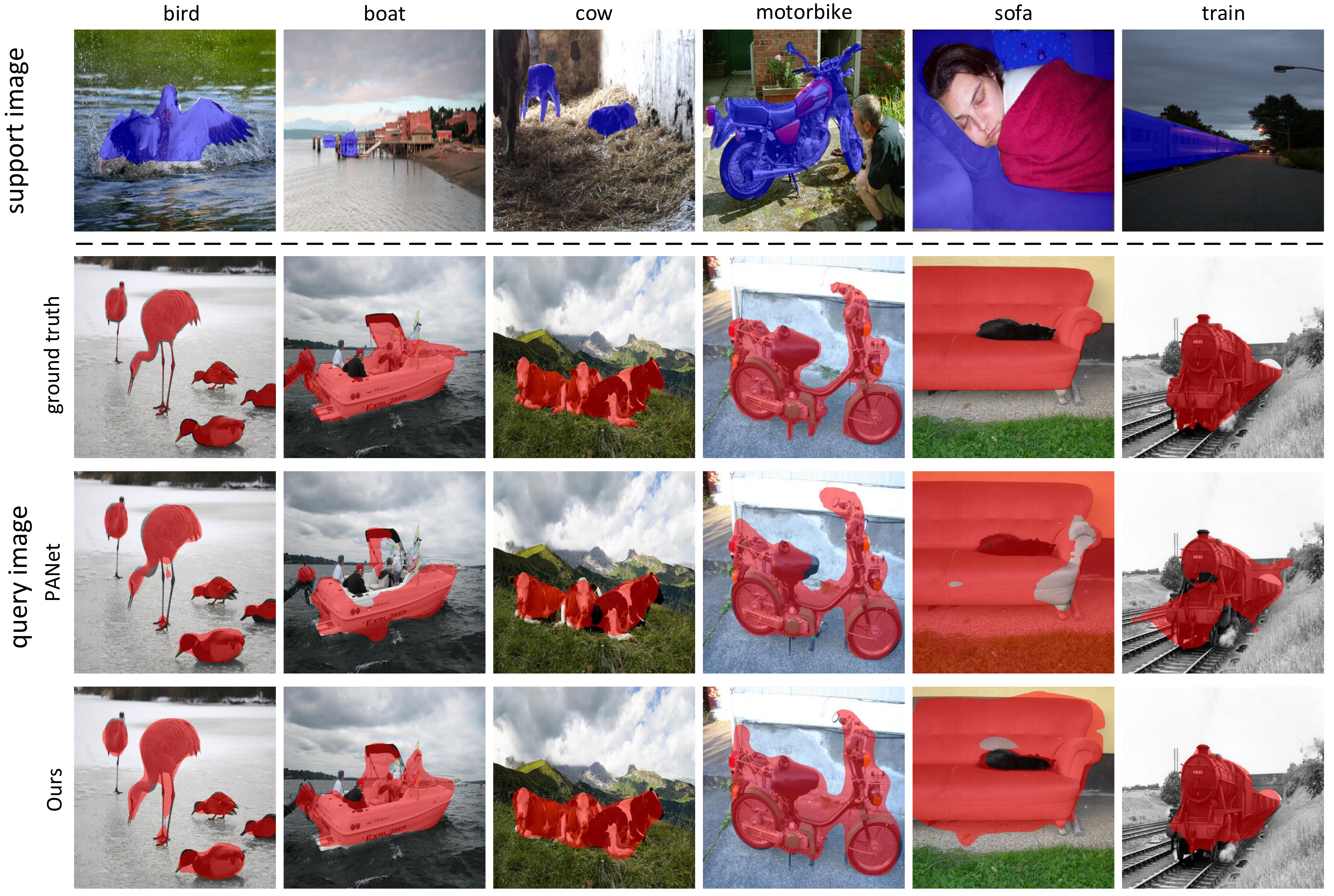}
	\caption{Example results of one-shot image segmentation on PASCAL-$5^{i}$ dataset. From top to down, we demonstrate the support image with the mask label that defines the target object class, the query image with ground truth label, the query image with the prediction of PANet, and the query image with the prediction of our method. Best viewed in color.}
	\label{fig_quality}
\end{figure*}

\begin{table*}
	\setlength{\tabcolsep}{7mm}
	\renewcommand{\arraystretch}{1.3}
	\caption{Results of one-shot segmentation on COCO-$20^{i}$ dataset evaluated with mean-IoU metric.}
	\label{tab_coco_m}
	\centering
	\begin{tabular}{rccccc}
		\toprule
		\textbf{Method\:\:\:}  &\textbf{COCO-}\bm{$20^{1}$}&\textbf{COCO-}\bm{$20^{2}$}&\textbf{COCO-}\bm{$20^{3}$}&\textbf{COCO-}\bm{$20^{4}$}&\textbf{Mean}\\
		\hline
		
		PANet \cite{wang2019panet} & -&-&-&-&20.9\\
		\hline
		\textbf{Ours}  & 32.2&23.5&19.6&19.0&\textbf{23.6}\\
		\bottomrule
	\end{tabular}
\end{table*}

\section{Experiments}
\label{experiments}

\subsection{Datasets and Evaluation Metrics}

\textbf{Datasets.}  We train and evaluate our proposed method for one-shot segmentation on the PASCAL-$5^{i}$ \cite{shaban2017one} and COCO-$20^{i}$ \cite{hu2019attention} datasets. The PASCAL-$5^{i}$ is built from the dataset of PASCAL VOC 2012 \cite{everingham2010pascal} which is expanded by SBD \cite{hariharan2011semantic}. The 20 semantic classes in PASCAL VOC are evenly divided into 4 splits, each containing 5 classes. The COCO-$20^{i}$ is created from the MS COCO dataset \cite{lin2014microsoft} that contains 80 foreground categories. Similarly, these categories are evenly divided into 4 splits, and each split contains 20 object categories. Experiments are done in a cross-validation manner. We train the model on the seen classes in 3 splits and test the model for the unseen classes in the rest split.\\
\textbf{Evaluation Metrics.}
Intersection-over-Union (IoU) is taken as the evaluation metric:
\begin{equation}
	IoU = \frac{TP}{TP+FP+FN}, 
\end{equation}
where $TP$, $FP$, and $FN$ are the numbers of true positive, false positive, and false negative pixels, respectively. We adopt the mean-IoU as the primary metric for the model evaluation. We first calculate a standard Intersection-over-Union (IoU) for each foreground class given the predicted masks of the split. We then average the class-wise IoU of all the 5 classes as the mean-IoU for this split. To compare our method with the early approaches, we also report the Binary-IoU that treats all object categories as one foreground class and averages the IoU of foreground and background.

\subsection{Implementation Details}

For the feature encoder, we adopt the VGG-16 model \cite{simonyan2014very} pre-trained on ImageNet \cite{deng2009imagenet} as our backbone. We remove the last two pooling layers to make the resolution of output feature maps effectively 1/8 times the input image size. To enlarge the receptive field, we apply Atrous Convolution \cite{chen2017deeplab} in conv5 layers with a rate of 2. The fully-connected layers are replaced by two $3\times3$ convolution layers with a dilated rate of 4. We utilize the Atrous Spatial Pyramid Pooling (ASPP) \cite{chen2017deeplab} as the classifier for both the multi-class label guidance branch and the foreground-background one-shot image segmentation. We employ an up-sampling layer along with the softmax output of the classifiers to match the size of the input image.

Following the setting in PANet \cite{wang2019panet}, input images are resized to (417,417) and augmented using random horizontal flipping. We also average the results from 5 runs with different random seeds on both the PASCAL-$5^{i}$ \cite{shaban2017one} and COCO-$20^{i}$ \cite{hu2019attention} datasets. Each run contains 1,000 episodes. We use SGD \cite{bottou2010large} as the optimizer. The momentum and weight decay of SGD are 0.9 and $10^{-4}$. The initial learning rate is set to $2.5 \times 10^{-4}$ and is decreased using the polynomial decay with a power of 0.9. We set batch size = 1 for training on PASCAL-$5^{i}$ and set batch size = 4 for training on MS COCO-$20^{i}$. We conduct the parameter search to choose best parameters for the framework. The detailed parameter analysis for $\lambda_{mcl}$, batch size and the initial learning rate can be found in Section \ref{subsec_pa}. 

\subsection{Baselines}

We compare our one-shot semantic segmentation method with the following state-of-the-art (SOTA) methods: 1-NN \cite{shaban2017one}, LogReg \cite{shaban2017one}, Siamese \cite{shaban2017one}, OSLSM \cite{shaban2017one}, FG-BG \cite{rakelly2018conditional}, Fine-tuning \cite{rakelly2018conditional}, co-FCN \cite{rakelly2018conditional},  PL \cite{dong2018few}, A-MCG \cite{hu2019attention}, SG-One \cite{zhang2020sg}, PANet \cite{wang2019panet}, AMP \cite{siam2019amp}.

\subsection{Experimental Results}

We compare our approach with state-of-the-art methods on both PASCAL-$5^{i}$ and COCO-$20^{i}$ datasets. Experimental results of one-shot segmentation on the PASCAL-$5^{i}$ dataset evaluated with mean-IoU metric are shown in Table \ref{tab_voc_m}. From Table \ref{tab_voc_m}, we can observe that our method achieves the best segmentation results compared to other state-of-the-art approaches on split-1, split-2, and split-4. Compared with other parametric methods \cite{shaban2017one,rakelly2018conditional,siam2019amp,zhang2020sg}, our method improves the average result of the 4 splits from 46.3\% to 52.6\% mean-IoU. Our approach also outperforms the non-parametric metric learning method PANet \cite{wang2019panet} by 4.5 \% mean-IoU. As shown in Table \ref{tab_voc_b}, when evaluated with the binary-IoU metric, our approach achieves the best average results of the 4 splits with 68.7\% binary-IoU. Our performance is 2.7\% higher than the second-best result reported by PANet \cite{wang2019panet}. 

The experimental results of one-shot segmentation on MS COCO-$20^{i}$ dataset using mean-IoU and binary-IoU metric are reported in Table \ref{tab_coco_m} and Table \ref{tab_coco_b}, respectively. Our method achieves better or comparable segmentation results compared to other state-of-the-art methods.  
Fig.~\ref{fig_quality} presents some qualitative segmentation examples for the one-shot task on the PASCAL-$5^{i}$ dataset. The first row gives the support image with the binary mask annotation that defines the target object class. The second row shows the query image with the ground truth label. The third and fourth rows show the query image with the prediction of PANet \cite{wang2019panet} and our method, respectively. It can be seen that our proposed approach successfully segments the objects from these query images.

\begin{table}[h]
	\renewcommand{\arraystretch}{1.3}
	\caption{Results of one-shot segmentation on COCO-$20^{i}$ dataset evaluated with the binary-IoU metric. The experiments are conducted on 4 splits and the average results are reported.}
	\label{tab_coco_b}
	\centering
	\setlength{\tabcolsep}{6mm}
	\begin{tabular}{rc}
		\toprule
		\textbf{Method\:\:\:\:}  &\textbf{binary-IoU}\\
		\hline		
		A-MCG \cite{hu2019attention}   & 52\\
		PANet \cite{wang2019panet} & \textbf{59.2}\\
		\hline
		\textbf{Ours}  & 58.7\\
		\bottomrule
	\end{tabular}
\end{table}

\section{Ablation Studies}
\label{ab_study}

\subsection{Element-Wise Component Analysis}

In this part, we demonstrate the contribution of each component of our training framework to the one-shot image segmentation task. The experimental results on PASCAL-$5^{i}$ are given in Table \ref{tab_ab_train}. Specifically, experiments are conducted on four splits, and we report the mean result of the 4 splits. The first row of Table \ref{tab_ab_train} shows the performance of our base network (described in Section \ref{sec_base_net}). By observing Table \ref{tab_ab_train}, we can notice that our approach improves the mean-IoU of the base network from 46.4\% to 49.7\% by introducing the multi-class label information. Furthermore, utilizing our proposed pyramid feature fusion module, we can have another 1.1\% mean-IoU performance gain. However, as shown in Table \ref{tab_ab_train}, if we do not leverage the multi-class label information, the pyramid feature fusion module does not improve the performance of the base network. On the contrary, it brings a 0.7\% mean-IoU performance drop. This highlights the importance of the introduction of multi-class label information, which can encourage the feature encoder to generate more discriminative features. We argue that less semantically meaningful features are easier to cause the over-fitting of subsequent fusion networks for the guided features. It will thus deteriorate the segmentation result. With our proposed self-prototype training branch, we can further improve the result to 51.3\% mean-IoU.

\begin{table}[t]
	\renewcommand{\arraystretch}{1.3}
	\caption{Element-wise component analysis for the training framework. The one-shot semantic segmentation results on PASCAL-$5^{i}$ dataset are evaluated with mean-IoU metric. The experiments are conducted on 4 splits and the average results are reported. MCL: Multi-Class Label Guidance, PFF: Pyramid Feature Fusion Module, SPT: Self-Prototype Training Branch.}
	\label{tab_ab_train}
	\setlength{\tabcolsep}{6mm}
	\centering
	\begin{tabular}{*{4}{c}}
		\toprule
		\textbf{MCL}       &\textbf{PFF}       &\textbf{SPT}       &\textbf{Mean}\\
		\hline
		&          &          &46.4\\
		&\checkmark&          &45.7\\
		\checkmark&          &          &49.7\\	
		\checkmark&\checkmark&          &50.8\\
		\checkmark&\checkmark&\checkmark&51.3\\		
		\bottomrule			
	\end{tabular}	
\end{table}

\begin{table}[t]
	\renewcommand\arraystretch{1.3}
	\caption{Ablation study for prototype fusion during testing. The results of one-shot semantic segmentation on the PASCAL-$5^{i}$ dataset are evaluated with mean-IoU metric. The experiments are conducted on 4 splits and the average results are reported.}
	\label{tab_ab_test}
	\setlength{\tabcolsep}{6mm}
	\centering
	\begin{tabular}{lc}
		\toprule
		\textbf{\:\:\:\:\:\:Method} &\textbf{Mean}\\
		\hline
		Support Prototype & 51.3\\
		Pseudo-Prototype  & 52.0\\
		Prototype Fusion  & 52.6\\	
		\bottomrule			
	\end{tabular}	
\end{table}

\begin{table}
	\renewcommand\arraystretch{1.3}
	\caption{The ablation studies for the influence of multi-class label guidance on the final performance of one-shot semantic segmentation on PASCAL-$5^{i}$ dataset. Results are evaluated with the mean-IoU metric. The experiments are conducted on 4 splits and the average results are reported. MCL: Multi-Class Label Guidance.}
	\label{tab_ab_freeze}
	\setlength{\tabcolsep}{6mm}
	\centering
	\begin{tabular}{lc}
		\toprule
		\textbf{\:\:\:\:Method} &\textbf{Mean}\\
		\hline
		Our w/ MCL  & 52.6\\
		Our w/o MCL & 49.5\\
		Frozen Encoder & 50.0\\
		\bottomrule			
	\end{tabular}	
\end{table}

\begin{figure}[t]
	\centering
	\includegraphics[width=0.45\textwidth]{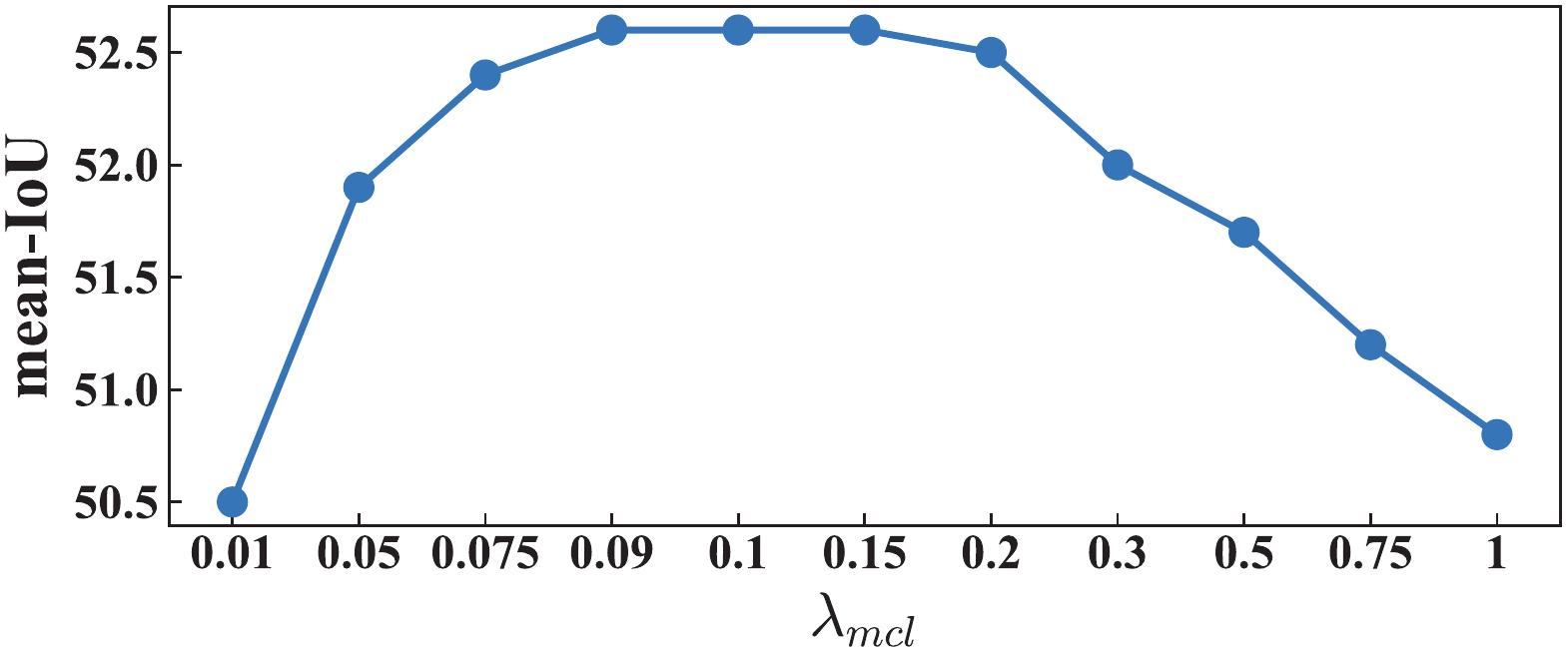}
	\caption{The parameter sensitivity of weight $\lambda_{mcl}$ for one-shot semantic image segmentation on PASCAL-$5^{i}$ dataset. Results are evaluated with the mean-IoU metric. The experiments are conducted on 4 splits and the average results are reported.}
	\label{fig_mcl}
\end{figure}

\begin{figure}[t]
	\centering
	\includegraphics[width=0.45\textwidth]{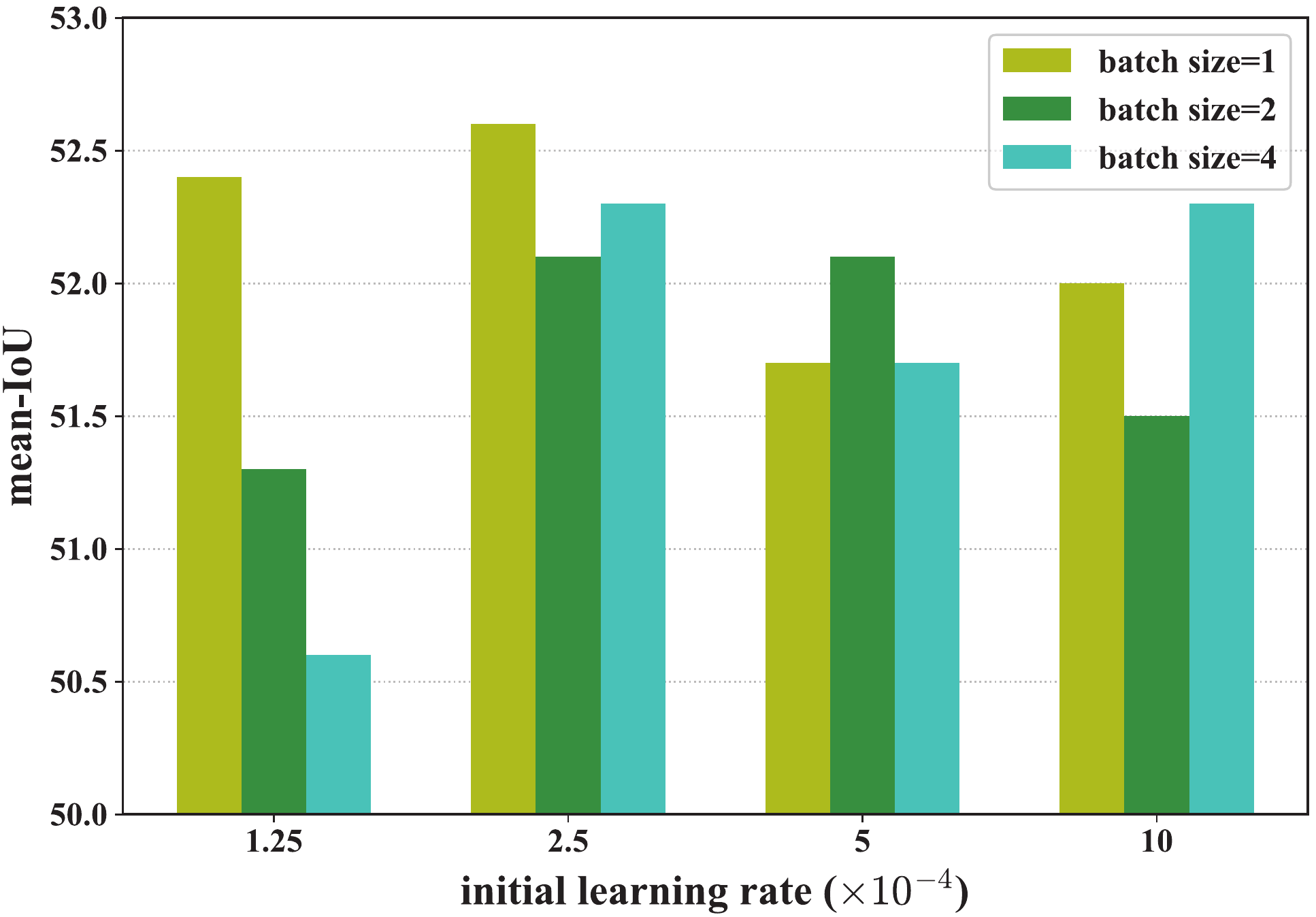}
	\caption{The parameter sensitivity of the initial learning rate and batch size for one-shot semantic image segmentation on PASCAL-$5^{i}$ dataset. Results are evaluated with the mean-IoU metric. The experiments are conducted on 4 splits and the average results are reported.}
	\label{fig_para}
\end{figure}

\subsection{Prototype Fusion for Testing} 
\label{subsec_ab}
In this part, we compare the performance of our prototype fusion strategy with that of using either the query pseudo-prototype or the support prototype only. Table \ref{tab_ab_test} gives the one-shot image segmentation results on PASCAL-$5^{i}$ dataset. As we can see, compared to using the support prototype, leveraging the pseudo-prototype of the query branch to guide the segmentation of the query image brings 0.7\% performance gain. The explanation is that due to the difference in the visual appearance of each image, the pseudo-prototype of the query image itself can provide more accurate guidance information of the target class for the query image, even if there exist noisy labels in the pseudo-mask. It should be noted that our self-prototype training branch also contributes to the improvement by adapting the network to fit the self-prototype guidance setting. Moreover, with our prototype fusion strategy, we further improve the mean-IoU to 52.6\%, which outperforms using either the query pseudo-prototype or the support prototype only.

\subsection{Influence of Multi-Class Label Guidance} 
As shown in Table \ref{tab_ab_freeze}, to further investigate the influence of the multi-class label guidance branch on the final segmentation results, we conduct experiments for the one-shot task on PASCAL-$5^{i}$ dataset when taking into consideration the following situations. \textbf{a}) Ours w/ MCL: Our full network with the multi-class label guidance branch. \textbf{b}) Ours w/o MCL: Our network without the multi-class label guidance branch.  \textbf{c}) Frozen Encoder: We train the network without the multi-class label guidance branch, and we also freeze parameters of the feature encoder. In other words, we directly load the ImageNet pre-trained weights to extract features for further processing. The last two pooling layers and full-connected layers are removed. From Table \ref{tab_ab_freeze}, we can notice that our proposed method with the multi-class label guidance branch brings the best segmentation performance. Without the multi-class label guidance, the performance of the network falls from 52.6\% to 49.5\%. We also notice that, without the multi-class label guidance, the performance of the network is even slightly worse than the result of loading the Imagenet pre-trained weights to extract features. We conjecture that the feature encoder trained without the multi-class label guidance generates less semantically meaningful features than the frozen encoder. This highlights the importance of leveraging the multi-class label guidance to extract the class-aware semantic feature representations for the one-shot semantic image segmentation.

\subsection{Parameter Analysis}
\label{subsec_pa}

For the parameter sensitivity of weight $\lambda_{mcl}$ in the multi-class label guidance branch, we evaluate the segmentation accuracy using the full proposed method on the PASCAL-$5^{i}$ dataset. We vary $\lambda_{mcl}$ over the range $\left \{0.01, 0.05, 0.075, 0.09, 0.1, 0.15, 0.2, 0.3, 0.5, 0.75, 1\right \}$. The mean results are reported in Fig.~\ref{fig_mcl}. As we can see, we get stable and better performance when $\lambda_{mcl}$ is between 0.075 and 0.2. We notice that a too small or large $\lambda_{mcl}$ can not facilitate the training process very much. According to the results, we empirically set $\lambda_{mcl}$ = 0.1 in our experiments.

For the parameter sensitivity of the initial learning rate and batch size, we conduct a parameter search to study their effects on the overall performance for one-shot semantic image segmentation. The experimental results on PASCAL-$5^{i}$ dataset are demonstrated in Fig.~\ref{fig_para}. As we can see, if the initial learning rate is relatively small, the better result is achieved when batch size = 1. If we increase the initial learning rate, we would better increase the batch size accordingly to get a better result. For example, it is suggested to set batch size as 2 and 4 when increasing the initial learning rate to $5 \times 10^{-4}$ and $10 \times 10^{-4}$, respectively. According to Fig.~\ref{fig_para}, we set the initial learning rate = $2.5 \times 10^{-4}$ and batch size = 1 for the best final result of 52.6\%.

\subsection{Speed Comparison}
In this part, we compare the speed of our approach with state-of-the-art methods. We reproduce the methods of SG-One \cite{zhang2020sg} and PANet \cite{wang2019panet} with their officially released codes. We run the one-shot segmentation task for 1,000 episodes during testing and report the average speed in Table \ref{tab_speed}. We also report the GPU and CPU consumption during testing for a fair comparison. As can be seen in Table \ref{tab_speed}, our approach can reach the speed of 3.5 FPS, which is comparable to the speed of PANet \cite{wang2019panet}. Though SG-One \cite{zhang2020sg} can reach the speed of 11.6 FPS, it requires much more computation resources, especially the CPU usage.

\begin{table}[t]
	\renewcommand{\arraystretch}{1.3}
	\caption{The Comparison of the speed for our approach with state-of-the-art methods on one-shot semantic segmentation.}
	\label{tab_speed}
	\centering
	\setlength{\tabcolsep}{3mm}
	\begin{tabular}{rccc}
		\toprule
		\textbf{Method\:\:\:\:}  &\textbf{GPU} &\textbf{CPU} &\textbf{FPS}\\
		\hline		
		SG-One \cite{zhang2020sg}  & 7.0G & 18$\times$&11.6\\
		PANet \cite{wang2019panet} & 1.8G &1$\times$&3.7\\
		\hline
		Ours  & 2.6G &1$\times$&3.5\\
		\bottomrule
	\end{tabular}
\end{table}

\subsection{Test with Weak Annotations}

\begin{figure}[t]
	\centering
	\includegraphics[width=0.47\textwidth]{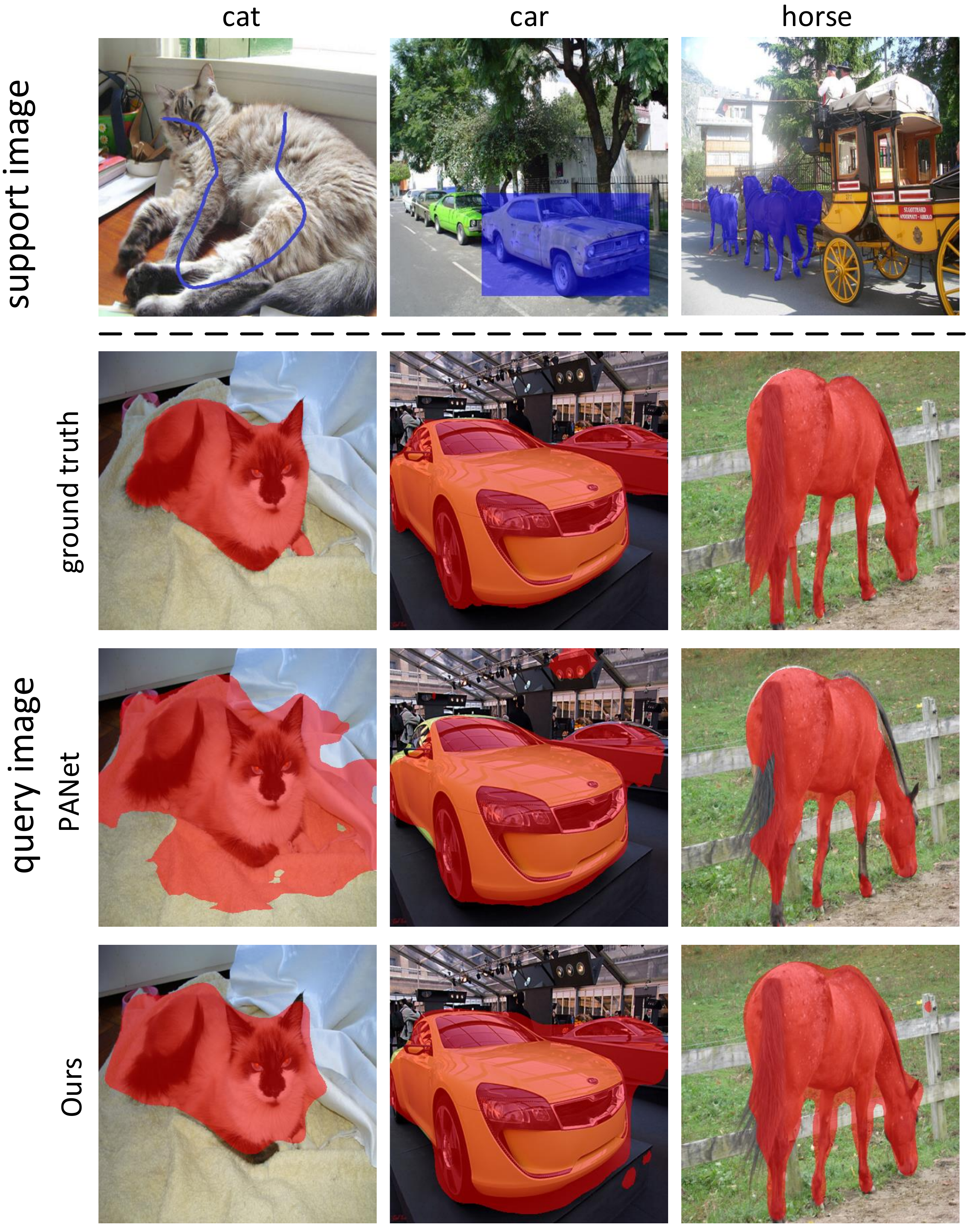}
	\caption{The qualitative results of our method on one-shot segmentation using scribble (left), bounding box (middle) and dense (right) annotations. The scribbles are dilated for better visualization. Best viewed in color.}
	\label{fig_weak}
\end{figure}

We further evaluate our model with the setting of weak annotations during testing. We replace the pixel-level dense annotations of the support set with scribbles or bounding boxes. Note that we only change the annotation setting of the support image during testing and use the same model trained with dense annotations. Following PANet \cite{wang2019panet}, these annotations are generated from the dense segmentation masks automatically. A random instance mask in each support image is chosen as the bounding box. The segmentation results on PASCAL-$5^{i}$ are given in Table \ref{tab_ab_weak}. We can notice that our method is much more robust than PANet \cite{wang2019panet}, especially for the test setting of scribble annotation. While our method only experiences a 0.7\% mean-IoU performance drop when changing dense annotation to scribble, the segmentation result of PANet \cite{wang2019panet} deteriorates from 48.1\% to 44.8\% seriously. Besides, our result of bounding box annotation is also very close to that of dense annotation. We notice that the performance of using scribble annotations is better than that of using bounding boxes in our experiment. The reason is that bounding boxes annotation will bring erroneous labels and make the support prototype less accurate, leading to a worse result than scribble annotation. Fig.~\ref{fig_weak} shows qualitative results of using scribble and bounding box annotations on the PASCAL-$5^{i}$ dataset. By observing Fig.~\ref{fig_weak}, we can find that even with weak annotations, our trained network can segment the objects successfully.

\begin{table}[t]
	\renewcommand\arraystretch{1.3}
	\caption{The performance of one-shot semantic segmentation with weak annotations during testing on PASCAL-$5^{i}$ dataset. Results are evaluated with the mean-IoU metric. The experiments are conducted on 4 splits and the average results are reported.}
	\label{tab_ab_weak}
	\setlength{\tabcolsep}{6mm}
	\centering
	\begin{tabular}{lcc}
		\toprule
		\textbf{Annotations}   & PANet \cite{wang2019panet} & \textbf{Ours} \\
		\hline
		\textbf{Dense}         & 48.1  & \textbf{52.6}\\
		\textbf{Scribble}      & 44.8  & \textbf{51.9}\\
		\textbf{Bounding box}  & 45.1  & \textbf{50.9}\\	
		\bottomrule			
	\end{tabular}	
\end{table}

\begin{figure}
	\centering
	\includegraphics[width=0.45\textwidth]{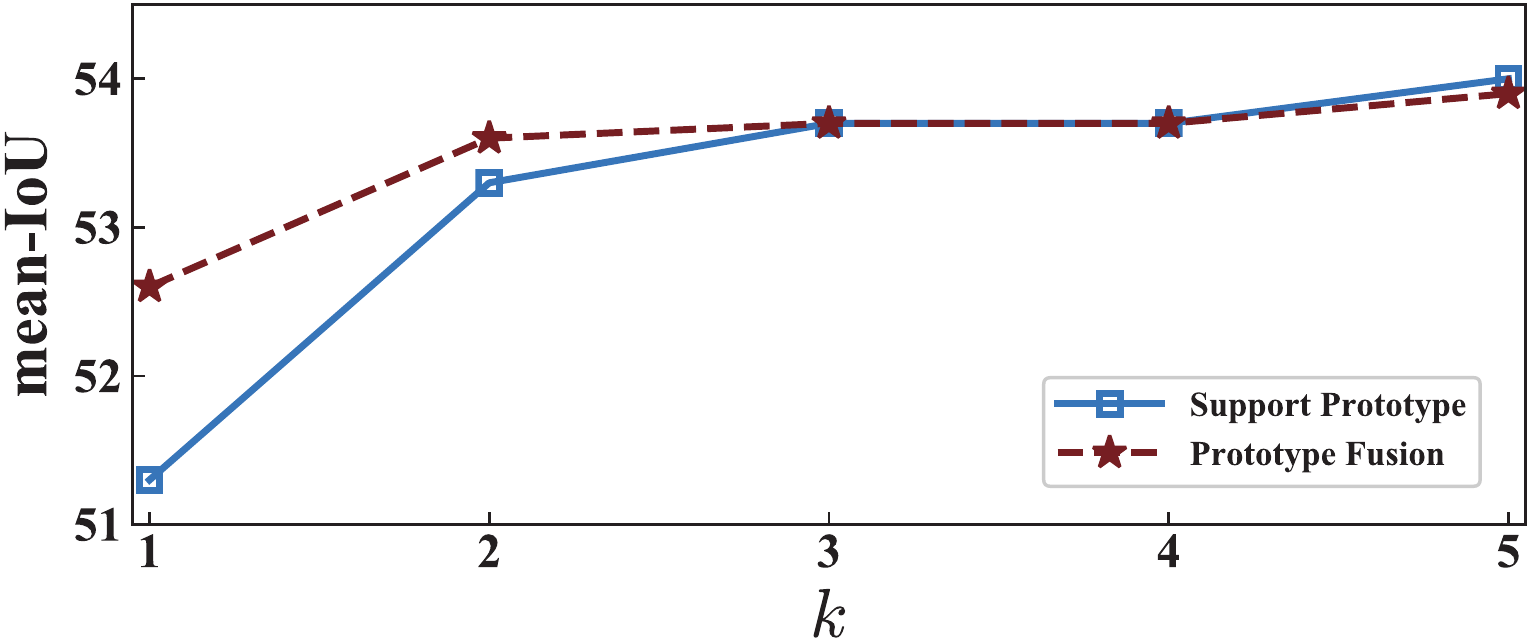}
	\caption{The results of k-shot segmentation on PASCAL-$5^{i}$ dataset evaluated with mean-IoU metric. The experiments are conducted on 4 splits and the average results are reported.}
	\label{fig_kshot}
\end{figure}

\subsection{Extension to K-shot Segmentation}

In the case of k-shot image segmentation, k labeled support images that contain the target object are provided to guide the segmentation of the query image. We directly use the trained network in the one-shot manner to test the segmentation performance by using k support images. We average the k target class prototype vectors extracted from each support image, and then use the averaged vector to guide the query image segmentation. In Fig.~\ref{fig_kshot}, we compare the result of averaging the k support prototypes only and the prototype fusion strategy that leverages the pseudo-prototype of query image (averaging k+1 prototypes). We notice that the fewer pictures, the more pronounced the advantages of the prototype fusion strategy. With more support images, the effect of the prototype fusion is gradually diluted. 

\subsection{Limitation}
In this part, we discuss the limitation of our proposed approach. From Fig.~\ref{fig_kshot}, we can find that more support images bring better segmentation results. However, the improvement encounters a bottleneck when the number of support images is more extensive than two. Therefore, it is essential to finetune the current model to take further advantage of more support images. In future work, we will study an efficient way to finetune the model when more support images are given. Various data augmentation strategies will also be investigated to facilitate the finetuning process.

\section{Conclusion}
\label{conclusion}

In this work, we proposed to leverage the multi-class label information to constrain the network during the episodic training of one-shot semantic image segmentation. Compared with existing methods, our proposed approach can help the network to extract class-aware semantic feature representations from the input images and thus generate a more semantically meaningful class prototype for the target category. In addition, we proposed to leverage a pyramid feature fusion module to mine the fused features for the target class in a multi-scale way. To take more advantage of the support image-mask pair, we further proposed a self-prototype guidance branch for support image segmentation. During testing, we proposed to combine the pseudo-prototype of the query image and the prototype from the support image to guide the final segmentation of the query image. Extensive experimental results on PASCAL-$5^{i}$ and COCO-$20^{i}$ datasets validated the superiority of our proposed approach.

\end{document}